# ADAPT: A Diffusion-based Adaptive Physics-aware Indoor Environmental World Model for Transferable HVAC Control

Xu Yang
Department of Automation
Tsinghua University, Beijing, China
yangxu24@mails.tsinghua.edu.cn

Kailai Sun*
SMART, Singapore
Massachusetts Institute of Technology
Cambridge, MA, USA
skl24@mit.edu

Dianyu Zhong
Department of Automation
Tsinghua University, Beijing, China

Qianchuan Zhao*
Department of Automation
Tsinghua University, Beijing, China
zhaoqc@tsinghua.edu.cn

## Abstract

Buildings account for about one-third of global energy consumption and $CO_2$ emissions. The optimization of indoor climate systems plays a critical role in urban climate mitigation, aligning with UN Sustainable Development Goals 11 and 13. However, building thermal dynamics are delayed and partially observable, while seasonal weather, solar radiation, and internal heat gains induce substantial differences across operating conditions. Existing methods struggle to accurately predict indoor thermal dynamics and generalize robustly across different conditions. To address these challenges, we propose ADAPT, a novel physics-aware conditional diffusion indoor environmental world model (IEWM) for accurate and robust indoor climate control. ADAPT leverages the expressive generative capability of diffusion models to predict a thermal baseline that explicitly captures latent thermal inertia. We further incorporate a multi-zone physical heat-balance regularization to constrain transferable thermal dynamics, enabling physically consistent thermal baseline generation without requiring manually calibrated thermal parameters. The learned IEWM is subsequently integrated into a downstream reinforcement learning controller for delayed credit assignment. Experiments on SemiBuildingSim and Sinergym environments demonstrate that introducing IEWM into indoor climate systems improves downstream control performance, reducing HVAC energy consumption by 7.3% and occupant discomfort by 30.2% compared with the strongest baseline. More importantly, under cross-season and cross-climate out-of-distribution transfer, ADAPTmaintains robust control performance with only marginal degradation relative to IID, substantially outperforming sequential and data-driven world models. This interdisciplinary study offers an energy-efficient, transferable AI solution for control science, building science, and energy sustainability. The code: https://github.com/xuyangthu88/ADAPT.

*Corresponding author.

## 1 Introduction

The United Nations reports that 55% of the world's population lived in urban areas in 2018, and this share is projected to reach 68% by 2050 [8, 11]. Cities concentrate energy use, emissions, infrastructure, and human exposure to climate risk [28, 40]. Buildings are a major part of this challenge. The UNEP and GlobalABC 2025 report further states that buildings consumed 32% of global energy and contributed 34% of global $CO_2$ emissions [5, 50]. Building energy efficiency is important for climate action and urban sustainability [4, 7].

Heating, ventilation, and air-conditioning (HVAC) systems contribute the largest share of building operational energy consumption [39]. At the same time, climate change is making heating and cooling demand increasingly dynamic, raising the requirements for closed-loop efficient HVAC control [51]. Recent studies suggest that accurately capturing weather-driven building thermal dynamics requires finer-grained temporal modeling, as evolving climate patterns introduce larger fluctuations to building thermal loads in many regions [46, 66, 67]. On the other hand, the HVAC system plays a key role as an interface between climate mitigation and human well-being. Indoor thermal comfort, as defined in ASHRAE Standard 55, is a basic condition for health, productivity, and occupant satisfaction [15, 47]. These global demands make occupant-centric HVAC control important at the intersection of energy systems and occupancy [48]. To support building decarbonization, intelligent HVAC control should capture critical weather-driven thermal dynamics and remain robust under changing weather and seasonal conditions, for energy efficiency and occupant comfort [21, 24, 55].

A practical challenge in HVAC control is the delayed and partially observed (PO) thermal response of buildings. Due to thermal inertia, heat is continuously stored and released by walls, floors, furniture, air volumes, and HVAC equipment, causing the effect of a control action to emerge only after multiple control intervals [53]. Meanwhile, occupancy, outdoor weather, and solar radiation jointly drive indoor thermal dynamics, while heat storage is not directly measured by sensors, rendering the latent thermal state only partially observable.

Model Predictive Control (MPC) has long been the classic framework for HVAC control [1, 14, 27]. However, MPC is sensitive to model mismatch, disturbance forecasts, and solver overhead in real deployments. More recently, Deep Reinforcement Learning (DRL) has emerged as a promising alternative by directly learning control policies from interaction data [32, 44, 52, 55, 56]. Nevertheless, most DRL methods remain fundamentally reactive and struggle to assign credit under delayed response systems [41]. Although recurrent architectures and Transformer-based policies partially mitigate delayed thermal responses and the PO problem by exploiting historical observations [36], they still implicitly infer latent thermal dynamics from past trajectories, resulting in limited generalization across different conditions.

These limitations naturally motivate learning an explicit predictive model of indoor thermal dynamics. Recently, world models have demonstrated remarkable success in long-horizon decision making by providing predictive representations for planning and value estimation in robotics and embodied AI [6, 13, 18, 19, 23, 58, 64]. Bringing this to HVAC control is appealing, as world model explicitly models delayed thermal dynamics and provides informative future trajectories for downstream controllers.

However, HVAC systems pose domain-specific challenges when applying world models in these systems. First, HVAC systems are partially observable and undergo substantial thermal distribution shifts across seasons, weather conditions, and climate regions, making accurate prediction considerably challenging. However, there is lack of world models that explicitly model indoor thermal dynamics for HVAC control. Second, sequential and vanilla data-driven world models tend to exploit specific statistical correlations of the training dataset rather than physically invariant thermal mechanisms. As weather, solar radiation, and climate shift out of the training distribution, their predictions become inaccurate and physically implausible, directly degrading downstream control performance.

To address the above challenges, we propose **ADAPT** (**A D**iffusion-based **A**daptive **P**hysics-aware indoor environmental world model for **T**ransferable HVAC control), a physics-aware conditional diffusion IEWM for robust HVAC control. Specifically, we introduce a generative conditional diffusion model to capture the distribution of future thermal baselines. To improve out-of-distribution generalization, we further introduce a physics-aware regularization that encourages generated baselines to satisfy fundamental thermal dynamics while preserving the ability of diffusion models. The regularization is based on a differentiable multi-zone physical heat-balance equation that models inter-zone exchange, outdoor-envelope exchange, solar radiation and internal heat gains. The IEWM is general and transferable, without requiring knowledge of the building geometry or hand-calibrated parameters. The predicted thermal baselines are used to support delayed credit assignment in downstream RL control. Our contributions are summarized as follows:

- We introduce ADAPT, a physics-aware conditional diffusion IEWM for HVAC control. The IEWM predicts the thermal baseline that captures delayed indoor thermal dynamics and support downstream RL credit assignment.
- We develop a physics-aware regularization based on a multi-zone heat-balance equation, encouraging diffusion-generated trajectories to satisfy fundamental thermal dynamics without requiring building-specific thermal parameters.
- We demonstrate that ADAPT consistently improves in-distribution control performance on SemiBuildingSim and Sinergym, reducing HVAC energy consumption by 7.3% and occupant discomfort by 30.2% compared with the strongest baseline.
- Extensive transfer experiments across seasons (summer↔winter) and climate regions (Stockholm↔Arizona) demonstrate that ADAPTachieves substantially stronger robustness than baselines under OOD scenarios, leading to more generalized downstream HVAC control.

## 2 Related Work

### 2.1 Reinforcement Learning and Transfer for HVAC Control

RL has become a leading data-driven alternative to model predictive control for HVAC operation, learning policies directly from interaction data without a hand-calibrated dynamics model [24, 32, 44, 54–56]. EnergyPlus-driven benchmarks such as Sinergym have further consolidated reproducible evaluation across zone temperature control and demand response [9]. Yet HVAC control remains structurally hard for model-free RL: hidden thermal storage introduces delayed action effects under partial observability, and standard temporal-difference learning routes credit through this delay only implicitly. Recent surveys emphasize this gap and the resulting brittleness across seasons and buildings [35, 38]. Transfer-oriented HVAC research has approached the problem by policy transfer across buildings [62] and online fine-tuning under distribution shift [12], but both still require target-domain interaction and offer limited guarantees when weather and internal loads move outside the training domain [2]. ADAPT takes a complementary route: rather than adapting the policy after deployment, it learns a generative indoor environmental world model whose predicted thermal baseline transfers across seasons and climate regions and is fed to a branch-wise action-value network [49], so the controller receives an explicit forecast of delayed thermal drift instead of an implicit latent state.

### 2.2 World Models for Decision Making

Learned dynamics have long been used for sample-efficient policy optimization and value expansion through PETS, MBPO, MBVE, and their extensions [16, 18, 26, 58, 63], and Dreamer, TD-MPC, and TD-MPC2 establish scalable latent world models for long-horizon planning [6, 19, 20, 22, 23]. Most of these methods target generic control domains where the observed state is largely sufficient and action-sequence search is the dominant challenge. HVAC control differs on both axes: observation is only a partial view of hidden thermal state, and the deployment distribution shifts across seasons and climate zones, so the world model is stressed more by *prediction* than by planning. ADAPT reflects this distinction by using a generative indoor environmental world model to forecast the future indoor trajectory under an imagined control assumption, and by providing the resulting thermal baseline to a downstream RL controller for baseline-conditioned action evaluation and delayed credit assignment.

### 2.3 Diffusion Models for Time Series and Control

Diffusion models generate data by reversing a noise process [17, 37, 45] and have been extended to probabilistic time-series forecasting and imputation [30, 42]. In sequential decision making, Diffuser recasts planning as trajectory denoising and diffusion policies generate control sequences [10, 25]; in energy, recent work applies diffusion to probabilistic building load forecasting [29, 57]. These works establish diffusion as a powerful tool for uncertain temporal prediction, but they do not target the two properties HVAC control needs most: physically consistent multi-step indoor trajectories under distribution shift, and a control interface that turns those trajectories into decision-relevant thermal baselines. ADAPTcloses this gap with a conditional diffusion IEWM whose multi-step rollouts are constrained by a learnable heat-balance regularizer, so generated trajectories remain on physically plausible manifolds across seasons, weather, and climate regions rather than reproducing source-domain correlations.

# 3 Method

We propose the ADAPT framework. The key idea is to learn a transferable indoor environmental world model (IEWM) that captures delayed indoor thermal dynamics, enabling HVAC controllers to make efficient decisions across various seasons and climate regions.

## 3.1 Problem Formulation

We model the optimal HVAC control problem as a Partially Observable Markov Decision Process (POMDP), defined by the tuple $\mathcal{M} = (\mathcal{S}, \mathcal{O}, \mathcal{A}, P, R, \gamma)$. In building environment, the true underlying physical state $\mathcal{S}$ includes unmeasurable thermal storage in structural mass (e.g., concrete walls and windows). However, at time step $t$, the agent only receives partial sensory observations $o_t \in \mathcal{O}$:

$$o_t = [\mathbf{T}_t,\, \mathbf{H}_t,\, \mathbf{O}_t^{occ},\, \mathbf{T}_t^{out},\, \mathbf{S}_t,\, \mathbf{Z}_t], \tag{1}$$

where $\mathbf{T}_t \in \mathbb{R}^N$ is the vector of zone temperatures, $H_t$ is relative humidity, $O_t^{occ}$ denotes measured or estimated occupancy pattern, $T_t^{out}$ denotes outdoor temperature, $S_t$ denotes solar radiation intensity, and $Z_t$ collects auxiliary sensor variables.

The controller maps observations to a multidimensional discrete action vector $a_t \in \mathcal{A}$, representing discretized HVAC control demands such as fan modes or temperature setpoints. The underlying thermal state evolves according to the thermodynamic transition $P(s_{t+1}|s_t, a_t)$. Due to the thermal inertia of building structures, the effect of an action at time $t$ may only become clearly visible after several control intervals. The objective is to optimize a policy $\pi$ that maximizes the discounted return: $\max_\pi \mathbb{E}_\pi \left[\sum_{t=0}^{\infty} \gamma^t r_t\right]$.

## 3.2 Generative IEWM Modeling

HVAC systems are partially observable and exhibit substantial distribution shifts across seasons, weather conditions, and climate regions, posing a greater challenge to world-model prediction than action-planning. To capture the resulting uncertainty and complex thermal dynamics, we formulate indoor environmental modeling as a conditional trajectory generation problem and learn a conditional diffusion IEWM, which models the distribution of a future thermal baseline conditioned on historical trajectories.

Specifically, the conditioning context consists of historical trajectories together with a held-action protocol that specifies the future control sequence, and is defined as

$$\mathbf{c}_t = (o_{t-N:t}, a_{t-N:t-1}, a_{t:t+H-1} = a_{t-1}), \tag{2}$$

By assuming held-action, the prediction trajectory $\hat{y}_t$ serves as a counterfactual baseline that captures the building's latent thermal inertia when the previous action $a_{t-1}$ is maintained. This baseline provides a reference for evaluating the long-term impact of the current control action, thereby improving delayed credit assignment in reinforcement learning. The IEWM predicts a future thermal baseline under the conditioning context:

$$\hat{\mathbf{y}}_t = \hat{o}_{t+1:t+H} \sim p_\theta(\mathbf{y}_t \mid \mathbf{c}_t), \tag{3}$$

To align the training objective, we enforce a held-action protocol ($a_{t:t+H-1} = a_{t-1}$) during offline data collection, utilizing the ground truth $o_{t+1:t+H}$ to provide supervision.

Given the conditioning context $\mathbf{c}_t$, the IEWM models the conditional distribution of future indoor thermal baseline using a denoising diffusion probabilistic model (DDPM). The diffusion process consists of a forward noising process and a learned reverse denoising process.

**Forward noising process.** Given a normalized future trajectory $\mathbf{y}_t^0$, the forward diffusion process gradually perturbs it with Gaussian noise. Following DDPM, the noisy trajectory at diffusion step $k$ is sampled directly from

$$q(\mathbf{y}_t^k \mid \mathbf{y}_t^0) = \mathcal{N}(\sqrt{\bar{\alpha}_k}\mathbf{y}_t^0,\, (1-\bar{\alpha}_k)\mathbf{I}), \tag{4}$$

where $\bar{\alpha}_k$ is a predefined variance schedule.

**Reverse denoising process.** The reverse process reconstructs clean trajectories conditioned on the HVAC context $\mathbf{c}_t$. Following the state diffusion formulation of GlobeDiff [61], the reverse Markov chain is defined as

$$p_\theta(\hat{\mathbf{y}}_t^{0:K} \mid \mathbf{c}_t) = p(\hat{\mathbf{y}}_t^K) \prod_{k=1}^{K} p_\theta(\hat{\mathbf{y}}_t^{k-1} \mid \hat{\mathbf{y}}_t^k, \mathbf{c}_t), \tag{5}$$

where the terminal distribution is $p(\hat{\mathbf{y}}_t^K) = \mathcal{N}(\mathbf{0}, \mathbf{I})$. Each reverse transition is parameterized as

$$p_\theta(\hat{\mathbf{y}}_t^{k-1} \mid \hat{\mathbf{y}}_t^k, \mathbf{c}_t) = \mathcal{N}(\hat{\boldsymbol{\mu}}_\theta(\hat{\mathbf{y}}_t^k, k, \mathbf{c}_t),\, \beta_k \mathbf{I}), \tag{6}$$

where the denoising network predicts the injected noise. The reverse mean is computed as

$$\hat{\boldsymbol{\mu}}_\theta = \frac{1}{\sqrt{\alpha_k}} \left( \hat{\mathbf{y}}_t^k - \frac{\beta_k}{\sqrt{1-\bar{\alpha}_k}} \hat{\epsilon}_\theta(\hat{\mathbf{y}}_t^k, k, \mathbf{c}_t) \right). \tag{7}$$

The denoising network $\hat{\epsilon}_\theta$ takes as input the noisy trajectory, the diffusion step, and the conditioning context, enabling trajectory generation consistent with the historical observations, weather conditions, and actions.

**Training objective.** For each training sample, we randomly select a diffusion step, inject Gaussian noise according to Eq. 4, and optimize the denoising network to predict the injected noise:

$$\mathcal{L}_{\text{diff}} = \mathbb{E}\left[ \left\| \epsilon - \hat{\epsilon}_\theta(\sqrt{\bar{\alpha}_k}\mathbf{y}_t^0 + \sqrt{1-\bar{\alpha}_k}\epsilon,\, k, \mathbf{c}_t) \right\|_1 \right]. \tag{8}$$

**Inference.** During inference, the model starts from Gaussian noise and iteratively performs reverse denoising conditioned on $\mathbf{c}_t$:

$$\hat{\mathbf{y}}_t^{k-1} = \hat{\boldsymbol{\mu}}_\theta(\hat{\mathbf{y}}_t^k, k, \mathbf{c}_t) + \sqrt{\beta_k}\epsilon_k, \qquad \epsilon_k \sim \mathcal{N}(\mathbf{0}, \mathbf{I}). \tag{9}$$

The final sample $\hat{\mathbf{y}}_t^0$ is taken as the predicted thermal baseline, denoted by $\hat{\mathbf{y}}_t$, which is used for evaluating delayed action effects.

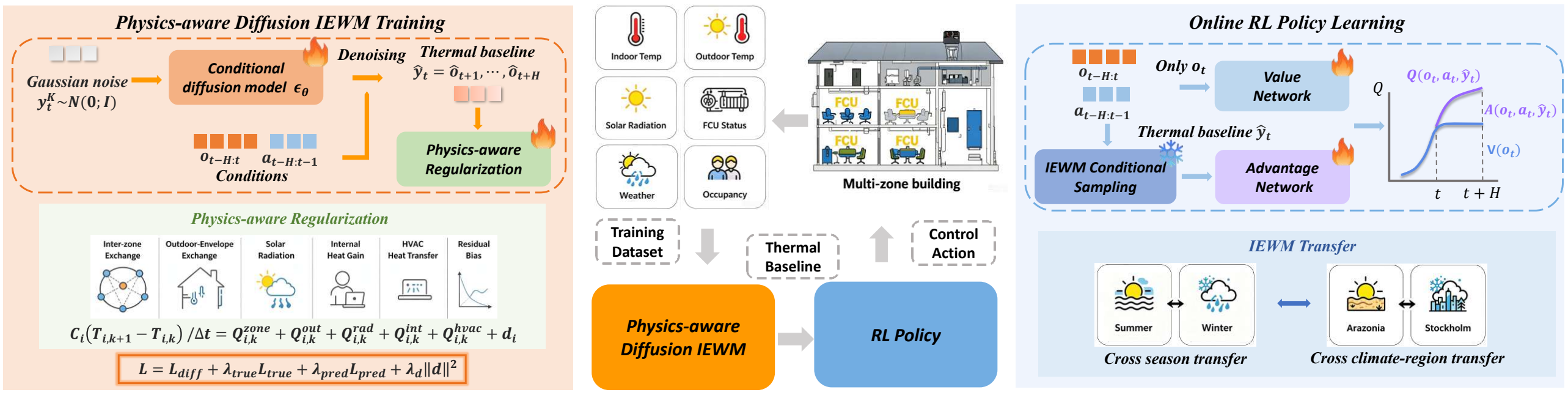

**Figure 1: Overall framework of ADAPT. The training process of ADAPT is divided into two phases. Phase I: Physics-aware Diffusion IEWM Training. Phase II: Online RL Policy Learning. IEWM is frozen during phase II.**

## 3.3 Physics-aware Thermal Regularization

The conditional diffusion IEWM can effectively model offline trajectories. However, vanilla data-driven learning may exploit domain-specific statistical correlations rather than invariant thermal dynamics. Such overfitting often generalizes poorly across seasons and climate regions. To improve out-of-distribution robustness, we introduce a *physics-aware regularization* based on a differentiable multi-zone physical heat-balance equation, which encourages generated trajectories to satisfy fundamental thermal dynamics while preserving the flexibility of the diffusion model.

For each zone $i$, the predicted temperature is encouraged to satisfy the following heat-balance equation:

$$C_i \frac{\hat{T}_{i,k+1} - \hat{T}_{i,k}}{\Delta t} = \hat{Q}^{zone}_{i,k} + \hat{Q}^{out}_{i,k} + \hat{Q}^{rad}_{i,k} + \hat{Q}^{int}_{i,k} + \hat{Q}^{hvac}_{i,k} + d_i. \quad (10)$$

Rather than serving as an exact physical constraint, Eq. (10) acts as a soft structural prior that regularizes diffusion predictions. All physical coefficients are jointly learnt with the diffusion model, allowing the regularizer to adapt to different buildings while maintaining physically meaningful heat-transfer mechanisms.

The heat-balance equation consists of five interpretable components. The inter-zone term models conductive heat exchange among neighboring zones using a symmetric nonnegative coupling matrix:

$$\hat{Q}^{zone}_{i,k} = \sum_{j=1}^{N} g_{ij}(\hat{T}_{j,k} - \hat{T}_{i,k}), \qquad g_{ij} = g_{ji} \geq 0. \quad (11)$$

The outdoor-envelope term captures heat transfer between indoor and outdoor environments:

$$\hat{Q}^{out}_{i,k} = k_i(T^{out}_k - \hat{T}_{i,k}), \qquad k_i \geq 0. \quad (12)$$

The radiation term models weather-dependent heat gains:

$$\hat{Q}^{rad}_{i,k} = \mathbf{s}_i^\top \mathbf{r}_k, \qquad \mathbf{s}_i \geq 0, \quad (13)$$

The internal-gain term represents heat generated by occupants:

$$\hat{Q}^{int}_{i,k} = \gamma_i \hat{o}^{int}_{i,k}, \qquad \gamma_i \geq 0. \quad (14)$$

Finally, the HVAC term models action-dependent heat exchange induced by the air-conditioning system, where $\hat{u}^{hvac}_{i,k} = \phi(a_{i,k})$ is learned:

$$\hat{Q}^{hvac}_{i,k} = \eta_i \hat{u}^{hvac}_{i,k}(T^{sup}_{i,k} - \hat{T}_{i,k}), \qquad \eta_i \geq 0. \quad (15)$$

A residual bias term $d_i$ captures varying unmodeled disturbances. We further penalize its magnitude to discourage the model from explaining missing physics through an unconstrained bias. Collectively, these structured components encourage the diffusion IEWM to capture transferable thermal dynamics instead of relying on specific data, thereby improving robustness in domain transfer.

**Joint Training with Data and Physics.** We jointly optimize the conditional diffusion IEWM and the physics-aware regularization through two complementary thermal residual objectives with distinct roles. The first residual is evaluated on ground-truth trajectories to identify the physical coefficients of the heat-balance model, independent of the diffusion model. Based on these learned coefficients, the second residual is evaluated on diffusion-generated trajectories to regularize the conditional diffusion IEWM, encouraging generated predictions to satisfy physical thermal dynamics.

For an observed trajectory, we define the ground-truth thermal residual as

$$\mathcal{R}_{i,k} = C_i \frac{T_{i,k+1} - T_{i,k}}{\Delta t} - (Q^{zone}_{i,k} + Q^{out}_{i,k} + Q^{rad}_{i,k} + Q^{int}_{i,k} + Q^{hvac}_{i,k} + d_i). \quad (16)$$

The corresponding parameter-identification loss is

$$\mathcal{L}_{\text{true}} = \frac{1}{HN} \sum_{k=t}^{t+H-1} \sum_{i=1}^{N} \rho(\mathcal{R}_{i,k}(\mathbf{y}_t)), \quad (17)$$

where $\rho(\cdot)$ denotes the Huber loss. Using the identified physical coefficients, we further define the thermal residual for diffusion-generated trajectories:

$$\hat{\mathcal{R}}_{i,k} = C_i \frac{\hat{T}_{i,k+1} - \hat{T}_{i,k}}{\Delta t} - (\hat{Q}^{zone}_{i,k} + \hat{Q}^{out}_{i,k} + \hat{Q}^{rad}_{i,k} + \hat{Q}^{int}_{i,k} + \hat{Q}^{hvac}_{i,k} + d_i). \quad (18)$$

The corresponding prediction regularization loss is

$$\mathcal{L}_{\text{pred}} = \frac{1}{HN} \sum_{k=t}^{t+H-1} \sum_{i=1}^{N} \rho(\hat{\mathcal{R}}_{i,k}(\hat{\mathbf{y}}_t)). \quad (19)$$

This objective directly regularizes the conditional diffusion IEWM. It encourages generated trajectories to satisfy the learned multi-zone heat-balance dynamics, thereby reducing physically implausible predictions and discouraging season-specific statistical overfitting under distribution shifts.

The overall training objective is

$$\mathcal{L} = \mathcal{L}_{\text{diff}} + \lambda_{\text{true}}\mathcal{L}_{\text{true}} + \lambda_{\text{pred}}\mathcal{L}_{\text{pred}} + \lambda_d \|\mathbf{d}\|_2^2. \quad (20)$$

where $\mathcal{L}_{\text{diff}}$ is the standard diffusion denoising objective, $\mathcal{L}_{\text{true}}$ identifies the parameters of the heat-balance model from observed trajectories, $\mathcal{L}_{\text{pred}}$ regularizes the conditional diffusion IEWM with physical dynamics. The bias penalty further discourages the regularizer from explaining missing physical effects through the unconstrained residual term, resulting in the IEWM that is both physically consistent and robust to out-of-distribution operating conditions.

### 3.4 Diffusion IEWM for RL learning

To leverage the proposed IEWM for downstream HVAC control, we integrate it into an online RL framework. After offline training, the IEWM is frozen and queried twice at each decision step. Before action selection, it predicts a thermal baseline, estimating how the indoor environment would evolve if the previous HVAC action were maintained. This baseline provides an explicit forecast of delayed thermal dynamics for action evaluation. After an action is selected, the same IEWM predicts a committed-action rollout by repeatedly applying the selected action, providing imagined future transitions for delayed credit assignment. The former improves action evaluation, whereas the latter improves temporal credit assignment.

For multidimensional HVAC control, we follow the branch-wise action-value architecture in [49], where each branch independently selects one actuator value $a_{t,d}$. The thermal baseline $\hat{y}_t$ is incorporated into the branch-wise action-value function as

$$Q_d(o_t, a_{t,d}, \hat{\mathrm{y}}_t; \Phi) = V_\psi(o_t) + A_{\eta_d}(o_t, a_{t,d}, \hat{\mathrm{y}}_t) - \max_{a' \in \mathcal{A}_d} A_{\eta_d}(o_t, a', \hat{\mathrm{y}}_t), \tag{21}$$

The thermal baseline is injected only into the advantage stream, enabling action preferences to be evaluated with explicit knowledge of future thermal dynamics while keeping state-value estimation based solely on current observation.

After executing the selected action, the environment provides the real one-step transition $(x_{t+1}, r_t)$. Starting from this, the IEWM predicts the future trajectory assuming the selected action is continuously executed

$$\hat{\mathrm{y}}'_t = [\hat{o}^{(a_t)}_{t+2}, \ldots, \hat{o}^{(a_t)}_{t+H+1}] \sim p_\theta(\mathrm{y} \mid o_{t-L+1:t+1}, a_{t-L+1:t}, \tilde{a}_{t+1:t+H} = a_t). \tag{22}$$

This committed rollout estimates the delayed thermal effect of executing the chosen action. Future rewards are evaluated using the environment reward function without learning an additional reward model. For horizon $h$, the expanded return is

$$G^{(h)}_t(a_t) = r_t + \sum_{k=1}^{h-1} \gamma^k \hat{r}^{(a_t)}_{t+k} + \gamma^h V_{\bar{\psi}}(\tilde{x}^{(a_t)}_{t+h}), \tag{23}$$

where $\hat{r}^{(a_t)}_{t+k}$ denotes the imagined future reward computed from the committed rollout, and $V_{\bar{\psi}}$ is the value function of the target network. To balance short and long-term delayed effects, the multi-step returns are aggregated into a TD-$\lambda$ target:

$$\tau^\lambda_t = (1-\lambda) \sum_{h=1}^{H-1} \lambda^{h-1} G^{(h)}_t(a_t) + \lambda^{H-1} G^{(H)}_t(a_t). \tag{24}$$

This target propagates delayed comfort and energy feedback through the IEWM prediction, enabling more effective credit assignment under long thermal inertia.

The branch-wise Q-network is optimized by Eq. 25. Full pseudocode is provided in Algorithm A.

$$\mathcal{L}_{\mathrm{RL}}(\Phi) = \mathbb{E}_{\mathcal{D}_{\mathrm{rl}}} \left[ \frac{1}{D} \sum_{d=1}^{D} \left( \tau^\lambda_t - Q_d(o_t, a_{t,d}, \hat{\mathrm{y}}_t; \Phi) \right)^2 \right]. \tag{25}$$

## 4 Experiments and Results

Our empirical evaluation is designed to answer three research questions (RQs):

- **RQ1: World-model benefit.** To what extent does IEWM improve HVAC control performance compared to existing HVAC control methods?
- **RQ2: Physics-aware modeling.** How effectively does the proposed physics-aware regularization improve prediction performance and thermal physical consistency over vanilla data-driven IEWMs?
- **RQ3: Transferable world model for control.** How effectively does the proposed physics-aware IEWM improve the HVAC control transferability under seasonal and climate-region transfer?

### 4.1 Experimental Setup and Evaluation Metrics

**Environments.** We introduce **SemiBuildingSim**, a high-fidelity HVAC simulation benchmark derived from a real 7-zone commercial office building in Hebei, China [59, 65]. The simulator is calibrated using over 9,000 time-aligned operational records, including weather, occupancy, and HVAC control logs, thereby capturing realistic occupant-driven disturbances, and long-horizon thermal inertia. We evaluate *seasonal transfer* by training and testing across the building's **summer** and **winter** operating conditions. The control interval is 5 minutes.

To further evaluate cross-climate robustness, we employ **Sinergym** [9], an EnergyPlus-based open-source benchmark, using the `2ZoneDataCenterHVAC` environment. We perform *climate-region transfer* between **Stockholm, Sweden**, characterized by a cold continental climate, and **Arizona, USA**, characterized by a hot desert climate, creating a challenging out-of-distribution evaluation across different climatic conditions. The control interval is 15 minutes.

**Prediction metrics.** We report mean absolute error (MAE), root mean squared error (RMSE), coefficient of variation of the RMSE (CVRMSE) for key temperature and humidity dimensions, and occupancy exact-match rate (Occ EMR) when applicable. Room-temperature and return-temperature CVRMSE are the primary evaluation metrics, as they are the key thermal variables governed by the multi-zone RC heat-balance equations and directly reflect both physical consistency and control-oriented prediction quality.

**HVAC control Metrics.** HVAC control is a multi-objective optimization problem that requires balancing thermal regulation, energy efficiency, and control stability. Accordingly, we adopt evaluation metrics that are tailored to the objectives of each benchmark. Thermal performance is benchmark-dependent: for SemiBuildingSim, we report occupant-centric comfort metrics, whereas for the Sinergym, we report the average temperature violation.

- **Occupant-centric thermal comfort:** We evaluate thermal comfort using the Predicted Percentage of Dissatisfied (PPD) and the absolute Predicted Mean Vote (|PMV|) according to ASHRAE Standard 55. Comfort metrics are computed primarily during occupied periods.
- **Energy consumption:** We report the total HVAC electricity consumption (kWh), which quantifies the energy efficiency and operational cost of the control policy.
- **Actuator action fluctuation:** We evaluate control smoothness using $\mathrm{AF} = \frac{1}{T}\sum_{t=1}^{T}\sum_{d=1}^{D}(a_{t,d} - a_{t-1,d})^2$, where lower values indicate smoother actions, reducing unnecessary high-frequency control oscillations and improving operational stability.

- **Average temperature violation:** Defined as the time-averaged absolute temperature violation outside the prescribed target temperature range, where zero indicates that the indoor temperature always remains within the acceptable bounds.

## 4.2 Main Results

**Answer for RQ1:** To answer RQ1, we first evaluate ADAPT on SemiBuildingSim against MPC [22], model-free RL (A2C [33], DQN [34], PPO [43], BDQ [49]), history-aware RL (TransformerRL [36]), and model-based RL (MBVE [16], MBPO [26], DreamerV3 [20]). All controllers are trained and evaluated under the in-distribution (ID) setting. As shown in Tab. 4, ADAPT achieves the best balance between energy efficiency, occupant comfort, and control smoothness, which demonstrates that introducing the diffusion IEWM provides a clear benefit for HVAC control.

**Table 1: Performance comparisons on SemiBuildingSim Summer (mean ± std over three seeds, ID). Lower is better.**

| Algorithm | Energy (kWh) ↓ | Abs PMV ↓ | PPD (%) ↓ | Action Fluctuation ↓ |
|---|---|---|---|---|
| MPC | 182.34 ± 1.59 | 0.60 ± 0.04 | 17.45 ± 1.19 | 16.57 ± 0.93 |
| A2C | 157.81 ± 7.59 | 0.50 ± 0.06 | 14.05 ± 1.56 | 12.89 ± 3.35 |
| PPO | 154.83 ± 3.10 | 0.41 ± 0.01 | 12.37 ± 0.33 | 9.75 ± 0.85 |
| DQN | 155.77 ± 2.03 | 0.49 ± 0.06 | 13.19 ± 1.88 | 9.86 ± 1.52 |
| BDQ | 154.57 ± 5.23 | 0.37 ± 0.01 | 10.88 ± 0.78 | 10.20 ± 1.17 |
| TransformerRL | 149.34 ± 2.93 | 0.37 ± 0.03 | 10.05 ± 0.50 | 9.42 ± 1.05 |
| MBVE | 150.73 ± 5.88 | 0.36 ± 0.01 | 10.27 ± 0.14 | 9.61 ± 1.86 |
| MBPO | 150.99 ± 3.12 | 0.38 ± 0.04 | 10.47 ± 0.49 | 12.47 ± 1.85 |
| DreamerV3 | 149.43 ± 1.02 | 0.36 ± 0.01 | 9.88 ± 1.53 | 8.37 ± 0.15 |
| **ADAPT (Ours)** | **138.49 ± 1.71** | **0.22 ± 0.01** | **7.01 ± 0.05** | **6.73 ± 0.31** |

The improvement stems from the thermal baseline predicted by the IEWM, which summarizes the future evolution of the indoor thermal dynamics while maintaining the current action. By explicitly modeling delayed thermal responses and building thermal inertia, the controller can optimize against anticipated thermal dynamics rather than relying solely on instantaneous observations, thereby avoiding unnecessary heating and cooling actions.

To further understand this benefit, Fig. 2 compares the training curves of different methods. ADAPT converges substantially faster and reaches a higher return, indicating that the predicted thermal baseline significantly improves RL sample efficiency.

The Sinergym results (Fig. 3) further demonstrate that the benefits of the IEWM generalize beyond SemiBuildingSim. ADAPT simultaneously reduces temperature violation while increasing energy savings. The improvement is particularly pronounced during the summer months in Stockholm, when longer daylight hours and stronger solar heat gains produce larger and more dynamic cooling loads. By predicting a thermal baseline that captures delayed indoor thermal dynamics, the IEWM enables proactive cooling decisions, suppressing unnecessary HVAC actuation while maintaining temperatures within the desired operating range. Additional results are shown in C.

**Answer for RQ2:** To answer RQ2, we compare four IEWM designs: the full ADAPT (Ours), w/o Phys., which removes the proposed physics-aware regularization, MambaFormer [60], and a VAE-based [31] world model. Table 2 shows that the proposed physics-aware IEWM consistently achieves the best multi-step prediction performance under seasonal transfer. The improvement is particularly pronounced for room temperature and return temperature, while the purely data-driven diffusion model and other sequence modeling baselines degrade substantially under OOD conditions.

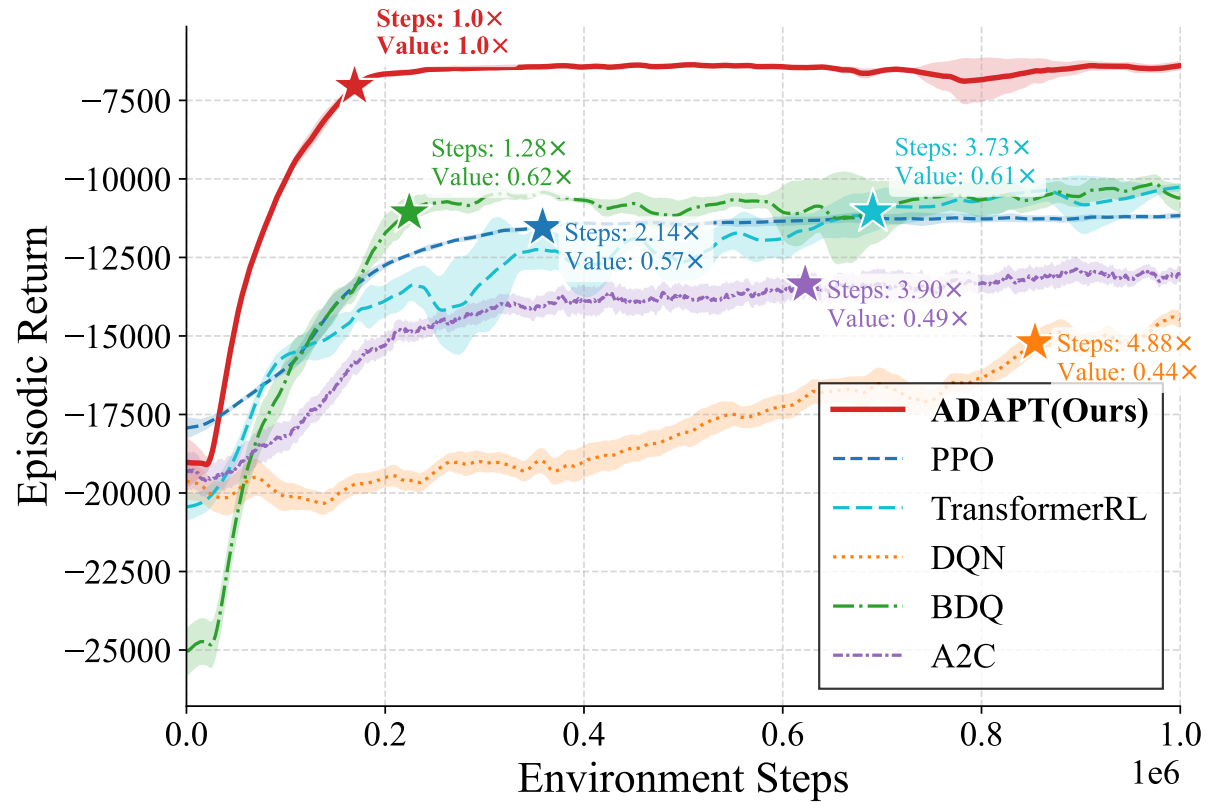

**Figure 2: Training curves on SemiBuildingSim Summer. Stars mark the first point where each method reaches 95% of its final episodic return. The labels report relative convergence steps (lower is better) and relative covergence return (higher is better), both normalized to ADAPT (Ours) = 1.0×.**

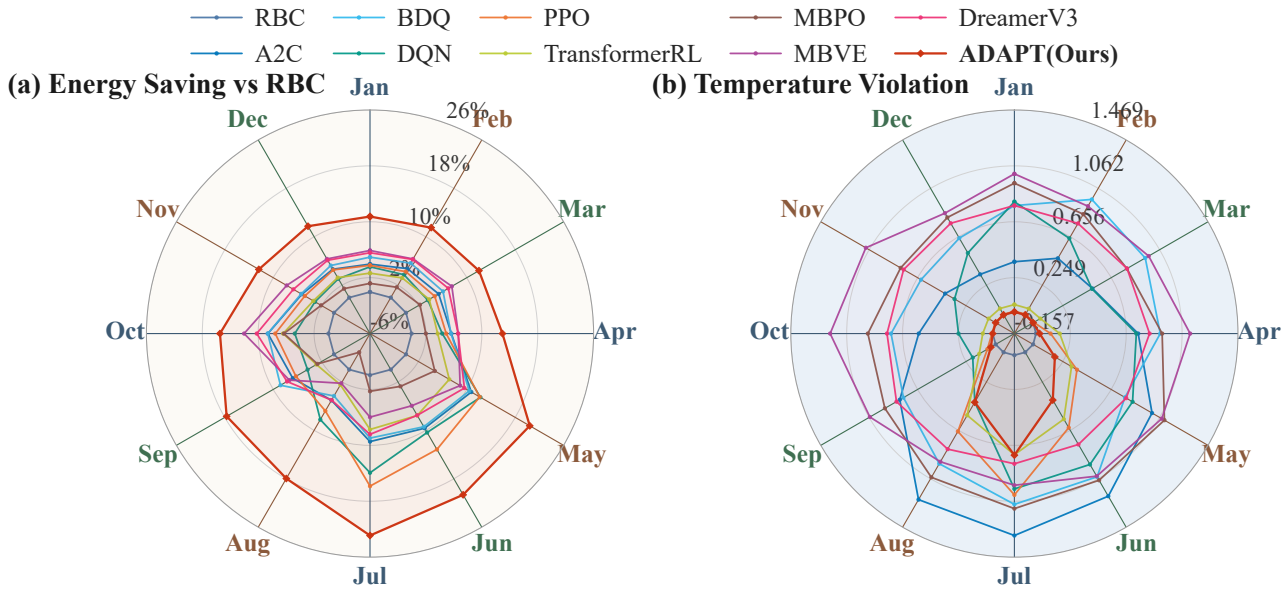

**Figure 3: Comparison of ADAPT with baselines in Sinergym Stockholm environment (ID). (a) energy savings relative to rule-based controller (RBC), (b) temperature violation.**

**Table 2: OOD prediction performance under seasonal transfer on SemiBuildingSim. Except Occ EMR, lower is better.**

| Category | Metric | ADAPT (Ours) | w/o Phys. | MambaFormer | VAE |
|---|---|---|---|---|---|
| **Summer → Winter (OOD)** | | | | | |
| Overall | MAE (↓) | **0.035** | 0.040 | 0.119 | 0.131 |
| | RMSE (↓) | **0.074** | 0.083 | 0.207 | 0.222 |
| | Occ EMR (%) (↑) | **98.475** | 98.086 | 77.021 | 82.855 |
| CVRMSE (%) | Return Temp | **4.633** | 8.069 | 9.517 | 13.367 |
| | Room Temp | **11.560** | 16.134 | 43.210 | 44.983 |
| **Winter → Summer (OOD)** | | | | | |
| Overall | MAE (↓) | **0.049** | 0.059 | 0.181 | 0.204 |
| | RMSE (↓) | **0.086** | 0.105 | 0.310 | 0.348 |
| | Occ EMR (%) (↑) | **77.486** | 74.892 | 76.977 | 75.538 |
| CVRMSE (%) | Return Temp | **6.531** | 9.237 | 18.105 | 21.279 |
| | Room Temp | **6.847** | 9.560 | 34.031 | 38.174 |

The gains are most evident on room temperature and return temperature because these variables are directly governed by the multi-zone RC heat-balance equations and therefore provide the most stringent test of physical consistency. By explicitly regularizing the denoising process with thermal dynamics, the proposed physics-aware objective prevents the diffusion model from exploiting season-specific statistical correlations and instead encourages predictions that remain consistent with the underlying heat-transfer process. Consequently, the learned world model maintains substantially better thermal-state prediction when weather conditions and building operating regimes change.

Figure 4 further demonstrates that the benefits of the proposed physics-aware IEWM generalize across climate regions. Under Stockholm and Arizona bi-transfer, ADAPT consistently achieves the best overall prediction performance across all reported metrics, whereas purely data-driven world models exhibit substantially larger degradation after climate transfer. These results demonstrate that incorporating physics-aware thermal constraints significantly improves the cross-region robustness and generalization ability of the learned indoor environment world model.

**Table 3: Ablation study on the physics loss weight $\lambda_{\text{pred}}$ and the bias penalty $\lambda_d$. Lower is better.**

| Ablation | Value | MAE ↓ | RMSE ↓ | CVRMSE (%) ↓ | Return Temp ↓ | Room Temp ↓ |
|---|---|---|---|---|---|---|
| $\lambda_{\text{pred}}$ | 0 | 0.059 | 0.105 | 13.671 | 9.237 | 9.560 |
| | 0.2 (Ours) | **0.049** | **0.086** | **10.513** | 6.531 | **6.847** |
| | 0.5 | 0.052 | 0.093 | 11.368 | 7.029 | 6.988 |
| | 0.8 | 0.056 | 0.101 | 12.501 | **6.314** | 8.321 |
| | 1.0 | 0.062 | 0.106 | 13.475 | 6.298 | 9.313 |
| $\lambda_d$ | 0 | 0.061 | 0.108 | 13.783 | 6.670 | 9.557 |
| | 0.1 (Ours) | **0.049** | **0.086** | **10.513** | 6.531 | **6.847** |
| | 0.3 | 0.051 | 0.088 | 10.868 | **6.143** | 6.935 |
| | 0.5 | 0.054 | 0.096 | 11.622 | 6.193 | 7.177 |

**Answer for RQ3:** To answer RQ3, we evaluate the downstream control performance of the IEWMs introduced in RQ2. To isolate the contribution of world-model design, all compared methods use the same RL controller described in Section 3.4, while only the underlying IEWM is varied.

Figure 5 shows that under seasonal transfer, ADAPTconsistently achieves the best trade-off between energy efficiency, occupant comfort, and control smoothness. The same trend is observed under climate-region transfer (Fig. 6). Across both transfer directions between Stockholm and Arizona, ADAPT consistently delivers higher energy savings with lower temperature violation than sequential and purely data-driven world models. The consistent advantage across different transfer scenarios demonstrates that improved world-model transferability leads to more robust and reliable downstream RL control. Additional results are shown in C.2.

The transferability originates from the proposed physics-aware regularization. By modeling climate-invariant thermal dynamics instead of environment-specific statistical correlations, ADAPT maintains accurate thermal prediction under OOD scenarios. Consequently, the downstream controller receives reliable future thermal baseline even in unseen environments, leading to more robust policy optimization and consistently stronger transfer performance.

**Ablation on Physics-Loss Weights.** Tables 3 investigates the influence of the two physics-aware regularization weights under the SemiBuildingSim Winter→Summer transfer. Removing either term consistently deteriorates OOD prediction performance, demonstrating that both physical consistency and residual-bias regularization are essential for learning transferable thermal dynamics. Conversely, excessively large weights also reduce prediction accuracy, indicating that overly strong physical constraints may compromise the expressive capacity of the diffusion model. We therefore use $\lambda_{\text{pred}} = 0.2$ and $\lambda_d = 0.1$ throughout the paper.

## 5 Discussion

As global urbanization and climate change reshape building energy demand, effective modeling and control of indoor thermal dynamics is important for building decarbonization, occupant comfort, and peak-load management, advancing the UN's SDGs 11 and 13. Our key contribution lies in introducing physics-aware generative indoor environmental world models as a transferable predictive foundation for HVAC control. Instead of learning a reactive controller tied to one season and one climate region, our framework utilizes world model to predict a thermal baseline that transfers across seasons and climate zones, supporting energy-efficient building operation, comfort improvement, and cross-region deployment for energy system and building science .

A central challenge in HVAC control is the combination of delayed thermal dynamics, partial observability, and substantial thermal distribution shifts across seasons, weather conditions, and climate regions. Since the effect of an HVAC action emerges gradually, effective control requires a world model that explicitly predicts future indoor thermal evolution for action evaluation and delayed credit assignment. We address this by introducing a physics-aware generative world model to guide RL learning. Besides, unlike existing deterministic and vanilla data-driven world models that primarily learn specific statistical correlations, our physics-aware diffusion IEWM learns transferable thermal dynamics through generative trajectory modeling regularized by a multi-zone physical heat-balance equation. Consequently, the proposed IEWM learns from one operating condition and is transferable across seasons, weather conditions, and climate regions.

In summary, our framework can offer building operators, HVAC engineers, and energy scientists a powerful generative tool for predicting delayed indoor thermal responses, transferring HVAC control across seasons and climate regions, and jointly improving energy consumption and occupant comfort across offices, residential buildings, and high-density AI infrastructure such as data-center cooling. This work established a decarbonized, physically consistent, and transferable HVAC control process aligned with SDGs 11 and 13, advancing the development of AI for building science, energy systems, and climate mitigation.

### 5.1 Limitations and Ethical Considerations

While the proposed physics-aware regularization significantly improves the transferability of the IEWM, it currently constrains only indoor temperature dynamics through the multi-zone heat-balance equation. More complex processes, such as humidity, are not explicitly modeled. Incorporating richer thermo-hygrometric dynamics into the world model is an important direction for future work.

All experiments are conducted using publicly available building simulators and weather data. No private, household-level, or personally identifiable information is collected, used, or inferred during model development or evaluation.

## 6 Conclusion

This paper presents ADAPT, a physics-aware diffusion IEWM for robust HVAC control under seasonal and climate-region transfer.

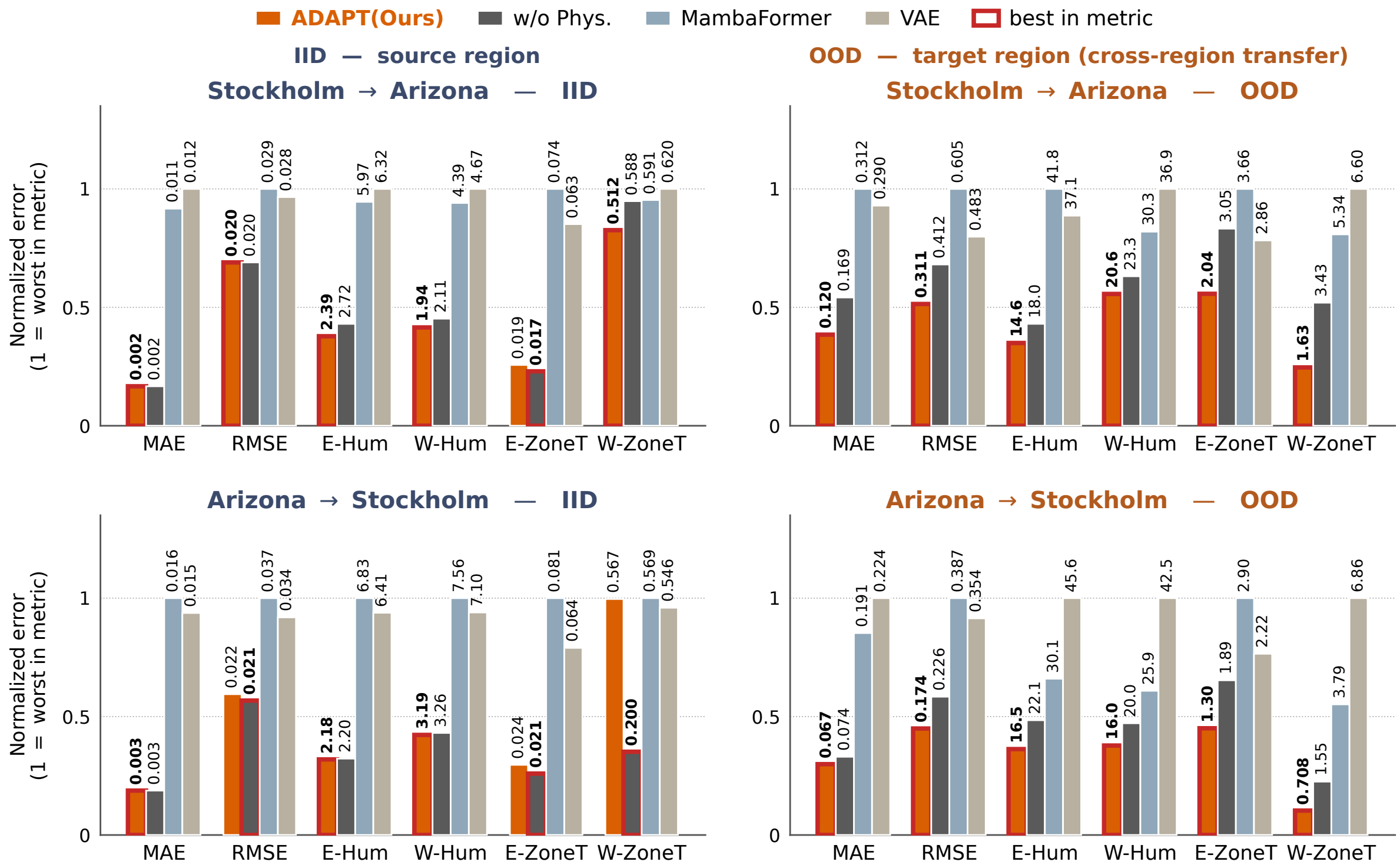

**Figure 4: Cross-region transfer prediction performance on Sinergym 2ZoneDataCenter (Stockholm ↔ Arizona). CVRMSE are annotated above each bar. Red outlines indicate the best value per metric (lower is better).**

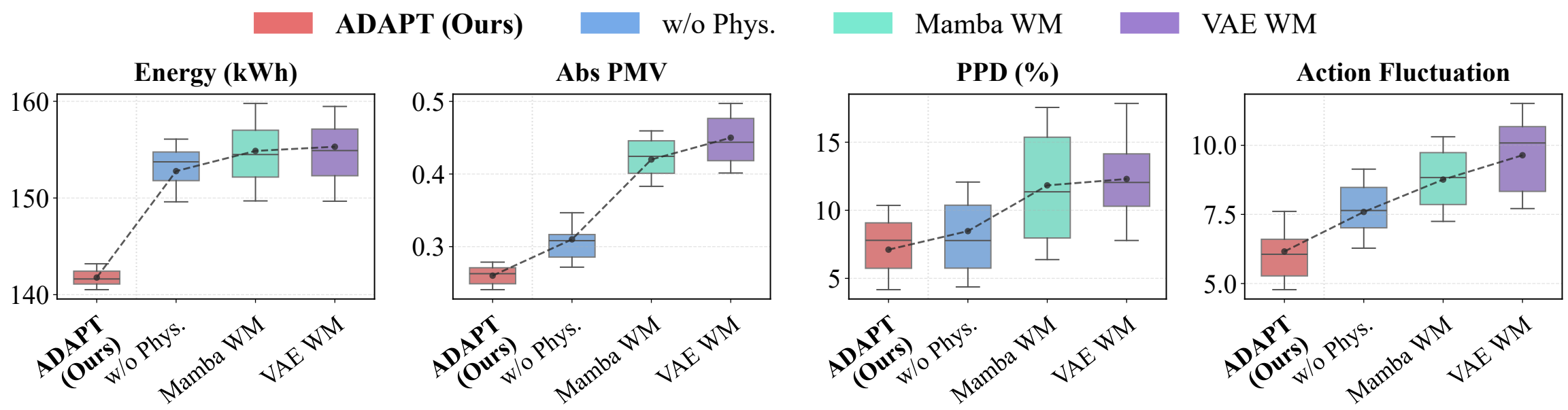

**Figure 5: Comparison of different IEWMs downstream control performance on SemibuildingSim Winter → Summer transfer. Lower is better.**

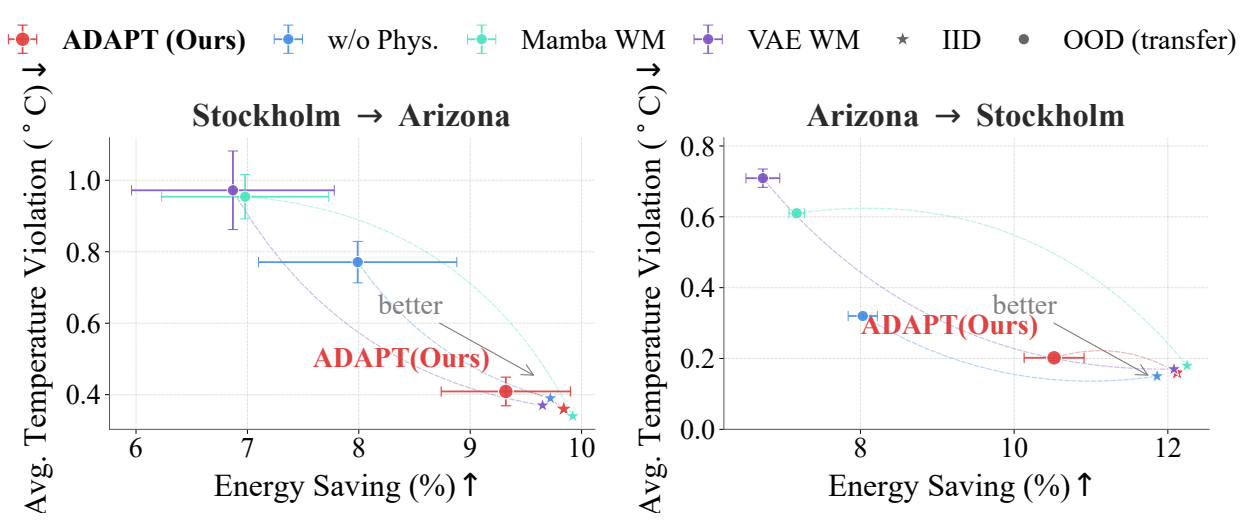

**Figure 6: Comparison of different IEWMs downstream control performance on Sinergym cross climate-region transfer.**

Experiments demonstrate that ADAPT consistently improves world-model prediction, downstream control performance, and transfer robustness over existing baselines. By integrating physics-aware thermal dynamics into generative world modeling, ADAPT provides a practical and transferable foundation for energy-efficient building control. We hope this work promotes further research on physics-guided world models for AI-driven building energy management, contributing to more sustainable and intelligent buildings.

## 7 GenAI Disclosure

The authors declare that AI tools were used only for language polishing and grammar checking. All scientific contributions, including ideas, experiments, and analyses, are the authors' own. All intellectual content, data analysis, interpretations, and conclusions were conceived, written, and verified by the authors.

# A Training and Inference Algorithm

Algorithm 1 summarizes the two-stage training and online-use procedure of ADAPT. Phase I trains the physics-aware diffusion IEWM by jointly minimizing the diffusion denoising loss (Eq. 8), the true-trajectory thermal residual (Eq. 17), the predicted-trajectory thermal residual (Eq. 19), and the residual-bias regularizer (Eq. 20). Phase II freezes the world model and runs baseline-conditioned RL: at each step, the frozen IEWM produces (i) the held-action thermal baseline $\hat{\mathbf{y}}_t$ that conditions the branch-wise action-value network (Eq. 21) and drives greedy action selection (Eq. **??**), and (ii) the action-committed rollout $\hat{\mathbf{y}}'_t$ that supplies delayed states for the held-action TD-$\lambda$ target (Eqs. 23–24). The policy parameters $\Phi$ are updated by minimizing $\mathcal{L}_{\text{RL}}$ (Eq. 25), followed by a soft update of the target network.

**Algorithm 1** Training and online use of ADAPT

**Require:** Offline trajectories $\mathcal{D}_{\text{off}}$, RL replay buffer $\mathcal{D}_{\text{rl}}$, horizon $H$, diffusion steps $K$, reward evaluator $g$, target update rate $\kappa$

**Phase I: Physics-aware diffusion IEWM training**

1: **for** each IEWM update **do**
2: Sample $(o_{t-N:t+H}, a_{t-N:t-1}) \sim \mathcal{D}_{\text{off}}$
3: Set $\tilde{a}_{t+h} \leftarrow a_{t-1}$ for $h = 0, \ldots, H-1$ and build $\mathbf{c}_t$ by Eq. 2
4: Set the clean target $\mathbf{y}_t^0 \leftarrow [o_{t+1}, \ldots, o_{t+H}]$
5: Sample $k \sim \text{Uniform}(\{1, \ldots, K\})$ and $\epsilon \sim \mathcal{N}(\mathbf{0}, \mathbf{I})$
6: Construct $\mathbf{y}_t^k \leftarrow \sqrt{\bar{\alpha}_k}\mathbf{y}_t^0 + \sqrt{1-\bar{\alpha}_k}\epsilon$ by Eq. 4
7: Predict $\hat{\epsilon}_\theta(\mathbf{y}_t^k, k, \mathbf{c}_t)$ and compute $\mathcal{L}_{\text{diff}}$ by Eq. 8
8: Generate $\hat{\mathbf{y}}_t$ and extract $\hat{\mathbf{T}}_{t+1:t+H}$
9: Compute $\mathcal{L}_{\text{true}}$ and $\mathcal{L}_{\text{pred}}$ by Eqs. 17 and 19
10: Update $\theta$ and thermal coefficients by minimizing Eq. 20
11: **end for**
12: Freeze $\theta$

**Phase II: Online RL Policy Learning**

13: **for** each environment step $t$ **do**
14: Generate $\hat{\mathbf{y}}_t$ from Eq. 9
15: **for** each branch $d = 1, \ldots, D$ **do**
16: Select $a_{t,d}$ with $Q_d(o_t, a, \hat{\mathbf{y}}_t; \Phi)$ from Eqs. 21
17: **end for**
18: Execute $a_t$; observe $o_{t+1}$ and $r_t$
19: Generate committed rollout $\hat{\mathbf{y}}'_t$ using Eq. 22
20: Store $(o_{t-L:t}, a_{t-L:t}, r_t, o_{t+1}, \hat{\mathbf{y}}_t, \hat{\mathbf{y}}'_t)$ in $\mathcal{D}_{\text{rl}}$
21: **if** it is time to update $\Phi$ **then**
22: Sample a minibatch from $\mathcal{D}_{\text{rl}}$
23: Compute $\hat{r}_{t+k}^{(a_t)}$ for $k = 1, \ldots, H-1$
24: Compute $G_t^{(h)}(a_t)$ and $y_t^\lambda$ by Eqs. 23 and 24
25: Update $\Phi$ by minimizing $\mathcal{L}_{\text{RL}}(\Phi)$ in Eq. 25
26: Soft-update target parameters $\bar{\Phi} \leftarrow \kappa\Phi + (1-\kappa)\bar{\Phi}$
27: **end if**
28: **end for**

# B Experimental and Environmental Details

## B.1 SemiBuildingSim Environment

*B.1.1 Environment Construction and Case Study Modeling.* To comprehensively evaluate the proposed control strategies, we utilize SemiBuildingSim, a high-fidelity simulator designed to replicate the complex thermal and energy dynamics of a real-world building. The simulation environment is grounded in historical data collected from a seven-zone office building in Hebei Province, China [59]. The building operates primarily during standard working hours (9:00-19:00 on weekdays), with reduced occupancy around lunchtime. An average of 12.4 people occupied the office (August 7 to August 22), with typical weekday usage and no regular weekend activity.

As shown in Fig. 7, the building comprises several individual rooms and a hall. One fan coil unit (FCU) is installed in each thermal zone. FCUs 1–4 each serve an individual cellular office, whereas FCUs 5–7 serve zones that allow inter-zone heat exchange. The FCUs receive chilled or heated water from a central refrigeration station comprising a heat pump and a circulating water pump. The pump distributes water to all FCUs through a closed-loop pipe network, while the heat pump provides the thermal source for cooling and heating. An overview of the HVAC system configuration and representative devices is shown in Fig. 8.

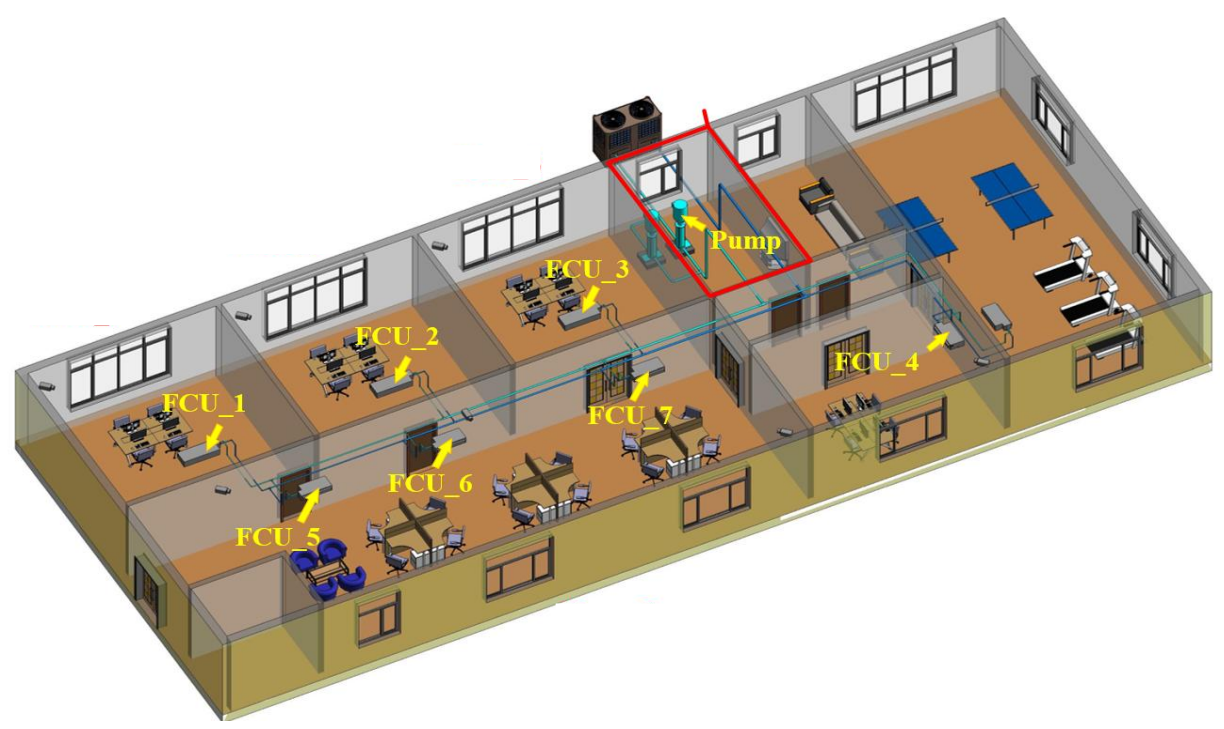

**Figure 7: Layout of the office building. Source from [65].**

The simulator comprises three rigorously coupled components: indoor zone models, parameterized HVAC equipment models, and a dynamic cooling water pipe network [59]. The indoor thermal dynamics are captured using a lumped capacitance approach that accounts for building envelope conduction (via the OTTV method), solar radiation, instantaneous occupancy loads, inter-zone heat transfer, and the sensible cooling/heating capacities delivered by the terminal HVAC units. The hydraulic pipe network utilizes a directed graph topology, solving the complete set of nonlinear hydraulic equations at each time step using the Newton-Raphson method to ensure mass continuity and system pressure-flow balance.

To support simulator calibration and monitor indoor thermal conditions, a comprehensive sensing infrastructure is deployed. At the zone level, each thermal zone is equipped with indoor air temperature sensors (Fig. 8(f)), and occupancy is inferred from ceiling-mounted video cameras (Fig. 8(c, g)). At the terminal side, each FCU is instrumented with a water flow meter, supply and return water temperature sensors, and pressure sensors (Fig. 8(b)). In the central plant, the variable frequency drive (VFD) circulating water pumps (1 Hz control resolution) are equipped with pressure and temperature sensors (Fig. 8(d)), and their electrical power is measured via electrical meters (Fig. 8(e)). All sensors operated continuously during HVAC runtime.

*B.1.2 Observation and Action Spaces.* The environment operates as a partially observable Markov decision process (POMDP), where the agent receives states and issues control commands at a fixed discrete timestep of $\Delta t = 5$ minutes.

**Observation Space:** The state vector $\mathbf{s}_t$ captures the current thermodynamic and operational context of the building. This includes the indoor air temperatures of the 7 zones $T_{\text{in},i}$, the instantaneous occupant counts $n_{\text{p},i}$, outdoor weather conditions, and the current operational states of the HVAC system components.

**Action Space:** The control variables actuate the terminal distribution systems across the building. The action space is formulated as a MultiDiscrete space, allowing for the independent control of the 7 FCUs installed in the respective thermal zones. For each FCU, the controller selects from 4 discrete fan operational modes (e.g., off, low, medium, high). Consequently, the joint action at time step $t$ is defined as $\mathbf{a}_t \in \{0, 1, 2, 3\}^7$.

*B.1.3 Performance Metrics and Reward Function.* To comprehensively assess the proposed building control strategies, we adopt a dual-objective evaluation framework that quantifies both occupant thermal comfort and HVAC energy use. These metrics define the reinforcement learning (RL) reward signal and are used for continuous controller performance evaluation.

*Thermal Discomfort and Energy Metrics.* Thermal comfort is evaluated using the Predicted Mean Vote (PMV) and Predicted Percentage of Dissatisfied (PPD) indices [15]. For the specific office conditions, the calculations assume an air velocity of $v = 0.15$ m/s, relative humidity of RH = 40%, clothing insulation of $I_{cl} = 0.63$ clo, and a metabolic rate of $M = 1.1$ met. Due to the well-insulated envelope, the mean radiant temperature $T_r$ is approximated to equal the zone air temperature $T_a$.

A PMV value of 0 indicates neutral thermal sensation, with an acceptable range of $[-0.5, +0.5]$ corresponding to a PPD below 10% [3]. The system-level discomfort at time step $t$ is computed as the occupancy-weighted mean PPD across all active zones:

$$\text{PPD}_{\text{mean},t} = \begin{cases} \frac{1}{N_t} \sum_{i=1}^{J} n_{i,t} \cdot \text{PPD}_{i,t}, & N_t > 0, \\ 0, & N_t = 0, \end{cases} \tag{26}$$

where $J = 7$ is the number of zones, $n_{i,t}$ is the occupancy count in zone $i$, and $N_t = \sum_{i=1}^{J} n_{i,t}$ is the total number of occupants.

Energy consumption is measured as the total HVAC electrical power $P_t$ (in kWh) at each time step $t$, returned by the simulator. Cumulative energy over an episode is computed as $E = \sum_t P_t \cdot \Delta t$.

*Composite Reward Function.* The objective of the RL agent is to maximize the expected cumulative reward. At each time step $t$, the composite reward $r_t$ is formulated to balance thermal comfort, energy conservation, and system operational stability:

$$r_t = -r_{\text{comfort}} - \alpha \cdot r_{\text{energy}} - \beta \cdot R_{\text{smooth},t} \tag{27}$$

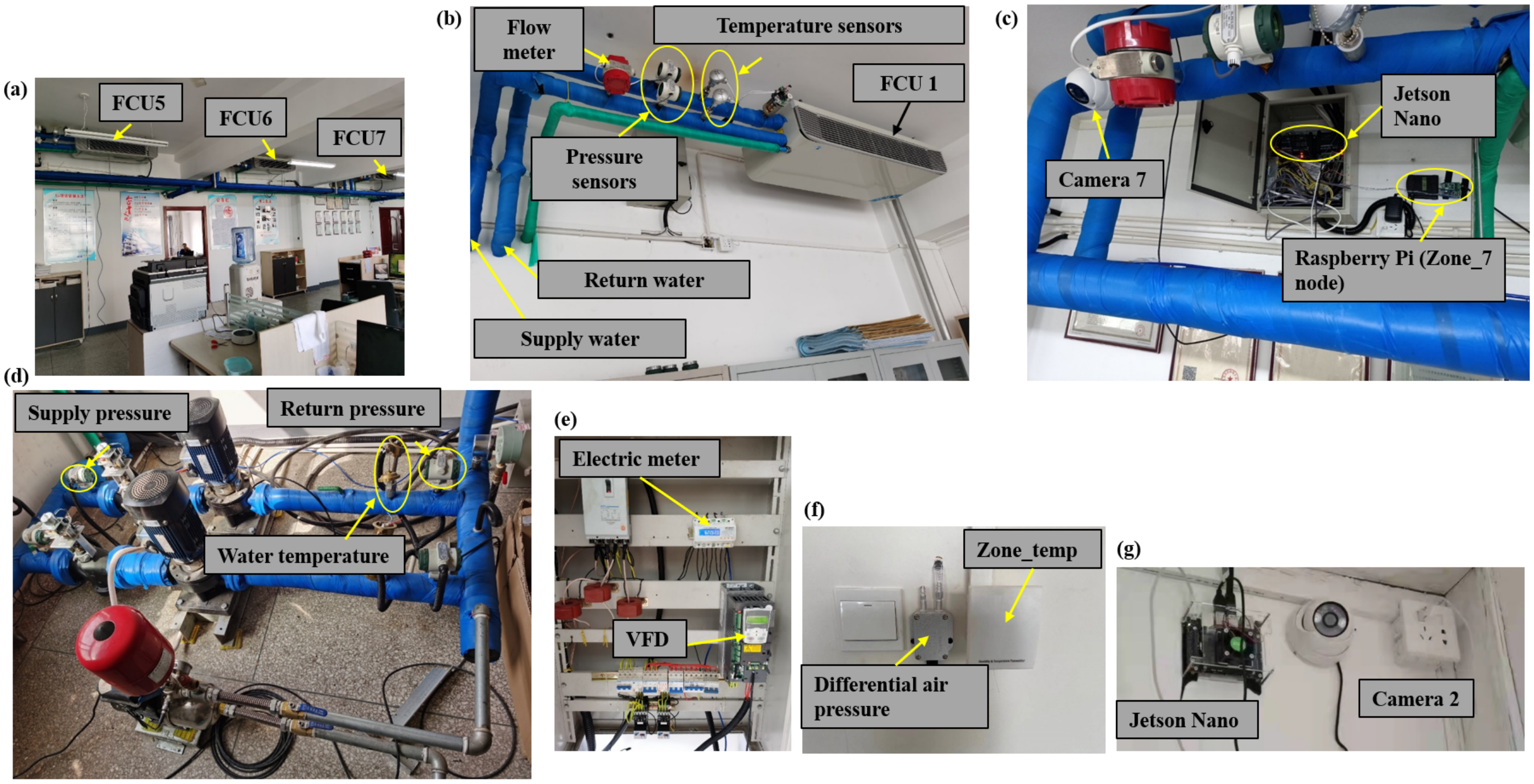

**Figure 8: HVAC system configuration. (a) FCUs of zones 5 to 7. (b) Sensors of FCU 1. (c) Video sampling of zone 7. (d) Sensors of water pumps. (e) Electrical cabinet. (f) Temperature sensor. (g) Video camera of zone 2. Source from [65].**

where $r_{\text{comfort}}$ penalizes thermal discomfort based on the calculated $\text{PPD}_{\text{mean},t}$, and $r_{\text{energy}}$ penalizes the total electrical power consumption $P_t$.

To optimize control performance and ensure physically plausible control, we introduce a smoothness penalty $R_{\text{smooth},t}$. This term explicitly discourages high-frequency oscillation and abrupt switching between FCU modes, thereby reducing mechanical wear and extending equipment longevity. The weighting coefficients are empirically established: the energy penalty coefficient is set to $\alpha = 20$. For the smoothness regularization, the coefficient is $\beta = 5$ when employing a Log smoothness reward, and $\beta = 3$ when utilizing a linear smoothness reward.

## B.2 Sinergym 2ZoneDataCenter Environment

*B.2.1 Environment Construction.* To benchmark our proposed control strategy against baselines, we utilize the `2ZoneDataCenter` environment provided by the open-source Sinergym framework [? ]. Sinergym serves as a Python-based virtual testbed that integrates deep reinforcement learning algorithms with the high-fidelity EnergyPlus building simulation engine.

The selected building model represents a single-story data center with a total surface area of 491.3 m$^2$. The facility is divided into two asymmetrical thermal zones: the West Zone and the East Zone. Unlike typical commercial buildings, the primary internal heat load in this environment is generated continuously by the hosted server racks (ITE objects), and the building has no windows or internal mass. To manage this intense thermal output, each zone is equipped with an HVAC system consisting of an air economizer, direct and indirect evaporative coolers, a single-speed direct expansion (DX) cooling coil, a chilled water coil, and a Variable Air Volume (VAV) system without reheat air terminal units.

*B.2.2 Observation and Action Spaces.* The environment is formulated as a discrete-time Markov Decision Process (MDP), where the agent interacts with the EnergyPlus backend via Sinergym's Gymnasium interface.

**Observation Space:** The state vector $\mathbf{s}_t$ encompasses both exogenous disturbances and the internal thermodynamic state of the data center. Key observations include outdoor weather variables (e.g., outdoor air temperature, relative humidity, solar radiation), the current indoor air temperatures of the West and East zones, and the instantaneous power consumption of the HVAC components and IT equipment.

**Action Space:** The control variables manipulate the thermal setpoints of the HVAC system to regulate the cooling capacity delivered to the servers. The action space dictates the continuous cooling and heating temperature setpoints for both the West and East zones. By dynamically adjusting these setpoints, the controller coordinates the mechanical cooling and economizers.

*B.2.3 Reward Function and Comfort Constraints.* In data center operations, maintaining a strict thermal environment is critical for equipment reliability, while minimizing energy consumption is essential for operational sustainability. Following Sinergym's official linear reward architecture, we define a dual-objective reward function that balances energy efficiency and thermal safety.

For our specific case study, the operational comfort temperature range for the server zones is strictly defined as $[T_{\text{low}}, T_{\text{high}}] =$ $[20°\text{C}, 26°\text{C}]$. The temperature deviation penalty $\Delta T_{i,t}$ at time step $t$ is calculated as the magnitude of the violation from this safe operating band for each zone:

$$\Delta T_{i,t} = \max(0, T_{i,t} - 26) + \max(0, 20 - T_{i,t}), \tag{28}$$

where $T_{i,t}$ is the indoor air temperature of zone $i \in \{\text{West}, \text{East}\}$.

The step reward $r_t$ computes a weighted sum of the energy penalty and the temperature violation penalty. To ensure both objectives are scaled appropriately, the terms are normalized relative to their maximum operational limits:

$$r_t = -\alpha \cdot \omega \cdot \left(\frac{E_t}{E_{\max}}\right) - \beta \cdot (1 - \omega) \cdot \left(\frac{\sum_i \Delta T_{i,t}}{T_{\max}}\right), \tag{29}$$

where $E_t$ is the total electrical power consumption of the HVAC system at time step $t$, and $E_{\max}$ and $T_{\max}$ are normalization constants established by the environment's empirical bounds. The weighting coefficient $\omega \in [0, 1]$ dictates the trade-off between energy conservation and thermal compliance. By imposing a hard penalty on temperatures outside the 20°C to 26°C threshold, the agent is incentivized to leverage free cooling dynamically without jeopardizing server integrity. We set $\omega = 0.5$, $\alpha = 1$, $\beta = 5 \times 10^{-3}$.

# C Additional Experiments

## C.1 Sinergym ID Control Comparison

We provide additional ID control results on two representative scenarios: the *Winter* setting of SemiBuildingSim and the *Arizona* setting of Sinergym. Tab. **??** and Fig. 9 show that ADAPT consistently achieves higher energy savings while maintaining occupant comfort within the desired temperature range. These results further demonstrate that the proposed physics-aware diffusion IEWM accurately captures building thermal dynamics under different seasons and climate conditions, providing reliable thermal baselines that consistently improve downstream HVAC control.

**Table 4: Performance comparisons on SemiBuildingSim Summer (mean ± std over three seeds, ID). Lower is better.**

| Algorithm | Energy (kWh) ↓ | Abs PMV ↓ | PPD (%) ↓ | Action Fluctuation ↓ |
|---|---|---|---|---|
| MPC | 251.00 ± 9.79 | 0.56 ± 0.06 | 16.20 ± 1.08 | 9.94 ± 1.13 |
| A2C | 247.28 ± 3.21 | 0.48 ± 0.03 | 13.05 ± 1.46 | 6.02 ± 1.03 |
| PPO | 236.52 ± 2.90 | 0.46 ± 0.04 | 12.62 ± 0.75 | 4.94 ± 1.46 |
| DQN | 248.41 ± 3.89 | 0.50 ± 0.05 | 13.02 ± 0.77 | 6.59 ± 0.55 |
| BDQ | 241.87 ± 2.73 | 0.45 ± 0.05 | 11.40 ± 1.28 | 5.40 ± 0.46 |
| TransformerRL | 233.97 ± 4.36 | 0.42 ± 0.03 | 11.72 ± 1.62 | 4.66 ± 0.65 |
| MBVE | 236.25 ± 3.03 | 0.42 ± 0.02 | 11.10 ± 0.18 | 5.08 ± 0.71 |
| MBPO | 234.66 ± 2.49 | 0.41 ± 0.01 | 11.34 ± 0.12 | 4.23 ± 0.77 |
| DreamerV3 | 228.84 ± 2.43 | 0.40 ± 0.02 | 10.65 ± 0.35 | 5.29 ± 0.88 |
| **ADAPT (Ours)** | **219.48 ± 1.50** | **0.28 ± 0.01** | **7.56 ± 0.05** | **3.62 ± 0.48** |

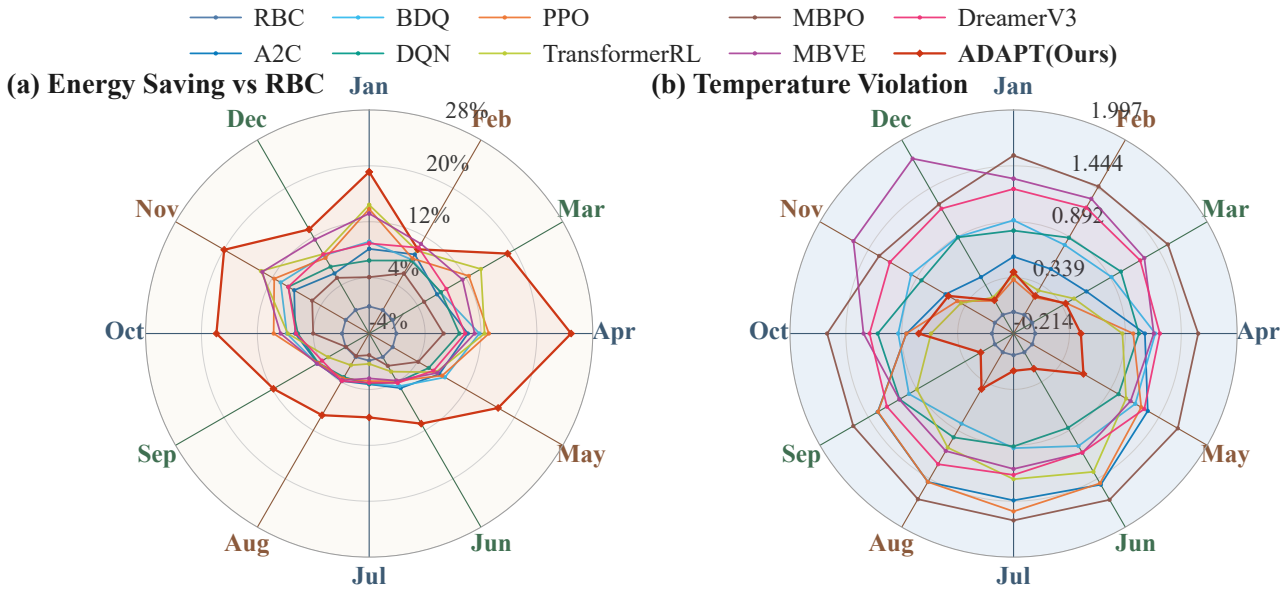

**Figure 9: Comparison of ADAPT with baseline controllers in Sinergym under the ID control setting for the Arizona environment. Subfigure (a) reports energy saving relative to RBC, and subfigure (b) reports temperature violation. Higher energy saving and lower temperature violation indicate better overall control performance.**

## C.2 SemibuildSim OOD Control Comparison

Figure 10 shows that under Summer→Winter seasonal transfer, ADAPTconsistently achieves the best balance between HVAC energy efficiency and occupant comfort. The proposed physics-aware diffusion IEWM accurately captures transferable building thermal dynamics across seasons, enabling robust downstream control with lower energy consumption while maintaining indoor temperatures within the desired comfort range.

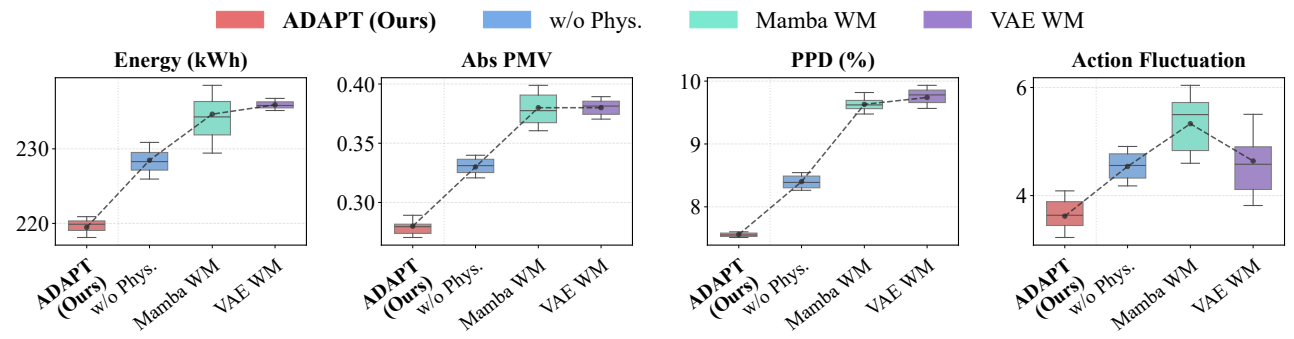

**Figure 10: Comparison of different IEWMs downstream control performance on SemibuildingSim Winter → Summer transfer. Lower is better.**

# D Implementation Details

## D.1 Implementation of World Models

This section describes the implementation details of the world models used in our experiments. The proposed method is **ADAPT** , the two neural world-model baselines are the MambaFormer and the VAE. All models are implemented in PyTorch and optimized with Adam.

*Dataset collection and input layout.* We construct transition datasets with a held-action protocol. Each training sample contains a history window as the conditioning variable and a flattened future observation sequence as the prediction target,

$$c_t = [\mathbf{o}_{t-N+1:t}, \mathbf{a}_{t-N+1:t-1}, a_{t:t+H-1} = a_{t-1}], \qquad \mathbf{y} = \mathbf{o}_{t+1:t+H}. \tag{30}$$

The previous action $a_{t-1}$ is held fixed over the forecast horizon, matching the dataset generation procedure used by all world models. In SemiBuildingSim, the control interval is 5 minutes, the context length is $N = 6$, and the forecast horizon is $H = 3$. In Sinergym, the control interval is 15 minutes, the context length is $N = 4$, and the forecast horizon is $H = 2$.

*Normalization and training protocol.* Continuous inputs and targets are normalized before training. Unless otherwise specified, models use a random seed of 3407, test split 0.2, batch size 64, learning rate $10^{-4}$, weight decay $10^{-5}$, maximum 200 epochs, gradient clipping with norm 1.0, and early stopping on validation RMSE with patience 12.

*ADAPT diffusion world model.* ADAPT formulates world-model prediction as conditional denoising. A Temporal U-Net denoiser $f_\theta$ is trained inside a Gaussian diffusion model with 20 denoising steps, dimension multiplier (8), hidden dimension 256, exponential moving average decay 0.995, and gradient accumulation every 2 mini-batches. The model uses an L1 denoising loss.

Given the conditioning vector $c_t$, the reverse process iteratively samples

$$\hat{\mathbf{y}} \sim p_\theta(\mathbf{y} \mid \mathbf{c}), \tag{31}$$

and the resulting denormalized sequence is evaluated as the multi-step world-model forecast. At evaluation time we use the EMA denoiser and enable clipped denoising for stable conditional samples.

*Thermal physics regularization.* ADAPT augments the denoising objective with a differentiable multi-zone RC residual. For room $i$, the residual is based on

$$C_i \frac{dT_i}{dt} = -[K\mathbf{T}]_i + k_i^{\text{out}} T_{\text{out}} + b_i^{\text{hvac}} u_i + b_i^{\text{occ}} O_i + d_i, \tag{32}$$

where $K = L + \text{diag}(k^{\text{out}})$ and $L$ is a learned symmetric graph Laplacian. The HVAC driver uses the held FCU fan mode and supply-water temperature,

$$u_i = -\phi_i(\text{fan}_i)(T_i - T_i^{\text{sup}}), \tag{33}$$

where $\phi_i(\cdot)$ is a learned monotone map over the discrete fan modes. The physical penalty combines a Huber penalty on the predicted trajectory residual, a Huber penalty on the true trajectory residual for RC parameter calibration, and a small bias regularizer:

$$\mathcal{L}_{\text{phys}} = \lambda_{\text{pred}} \rho_\beta(r_{\text{pred}}) + \lambda_{\text{true}} \rho_\beta(r_{\text{true}}) + \lambda_{\text{d}} \|d\|_2^2. \tag{34}$$

**Table 5: Physical-loss hyper-parameters of ADAPT.**

| Parameter | Value |
|---|---|
| Predicted RC residual weight $\lambda_{\text{pred}}$ | 0.2 |
| True RC residual weight $\lambda_{\text{true}}$ | 0.5 |
| Bias penalty weight $\lambda_{\text{bias}}$ | $1\times10^{-1}$ |
| Time step $\Delta t$ | 1.0 |
| Huber threshold $\beta$ | 0.25 |

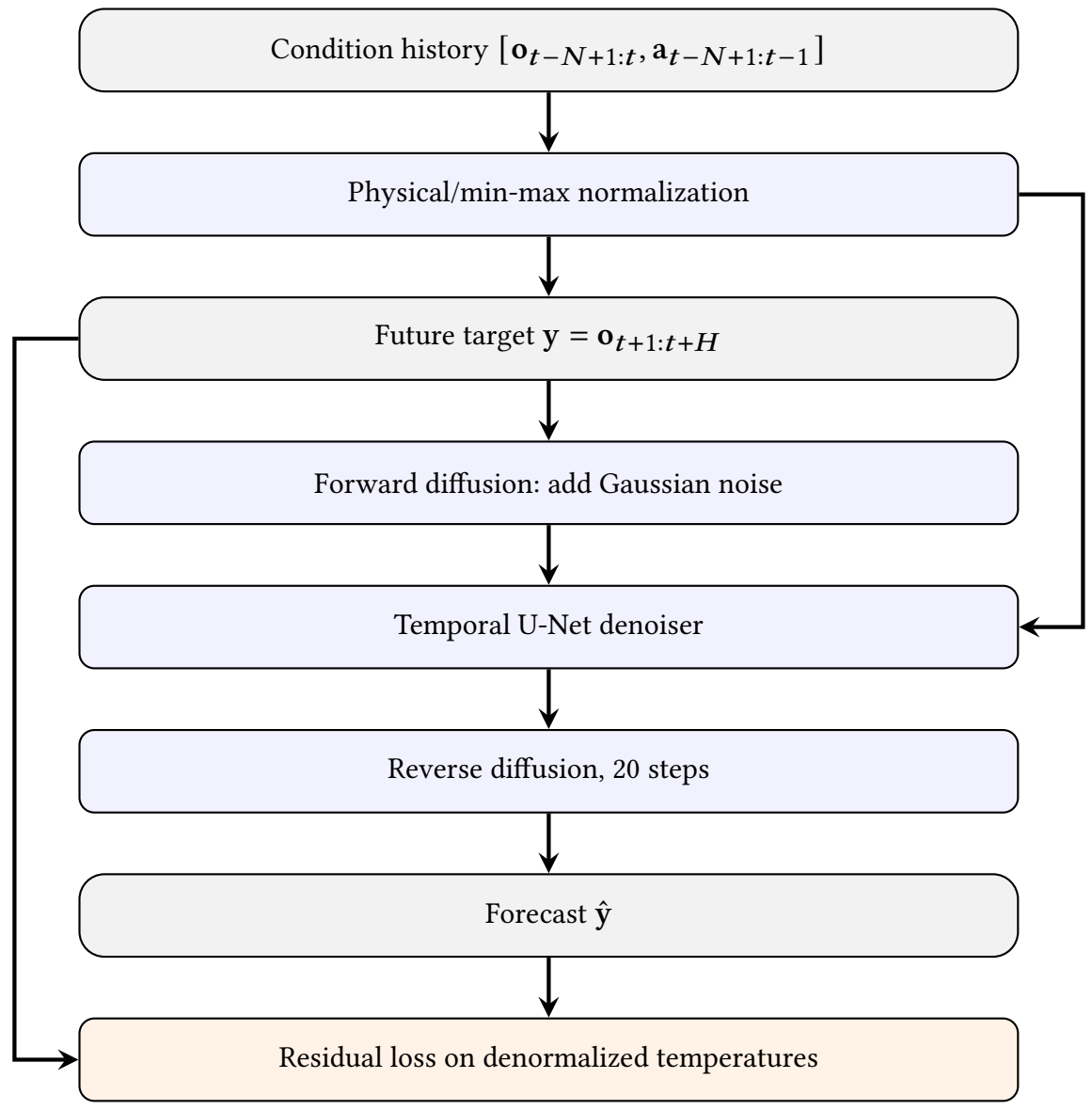

**Figure 11: ADAPT architecture for conditional diffusion world-model prediction.**

Tab. 5 reports the physical-loss parameters used by ADAPT .

*Evaluation metrics.* For continuous state variables, we report Mean Absolute Error (MAE), Root Mean Squared Error (RMSE), coefficient of variation of RMSE (CVRMSE), and $R^2$ where applicable. For SemiBuildingSim occupancy-related targets, we additionally report Occupancy Exact Match Rate (OCC EMR) after converting the normalized predictions back to the original scale.

*ADAPT architecture.* Figure 11 illustrates the ADAPT training and sampling pipeline. The conditioning history is repeated across the diffusion horizon, the Temporal U-Net denoises the future target vector, and the heat-balance equation regularizes the denormalized temperature trajectory during training.

## D.2 MPC Implementation Details

We use a sampling-based model predictive control (MPC) approach for online decision making. At each environment step, the controller plans over a finite horizon using the learned indoor environment world model (IEWM), evaluates imagined trajectories using a combination of a reward evaluator and a learned terminal value function, and executes only the first action of the best-scoring sequence. The procedure is repeated after the next observation is received.

*Planning objective.* Let $o_t$ denote the current observation and let $\mathbf{a}_{t:t+H-1} = (a_t, \ldots, a_{t+H-1})$ be a candidate action sequence of horizon $H$. The IEWM is used to recursively predict future states. We specifically utilize its single-step forward prediction,

$$\hat{o}_{t+k+1} = f_\theta(\hat{o}_{t+k}, a_{t+k}), \qquad \hat{o}_t = o_t, \tag{35}$$

where all planning rollouts are performed in the prediction space of the IEWM. To account for the long-term return beyond the finite planning horizon, we bootstrap using a learned terminal value function $V_\psi(\cdot)$. The quality of a candidate sequence is computed by accumulating discounted rewards and adding the temporal difference (TD) value estimate at the terminal state,

$$J(\mathbf{a}_{t:t+H-1}) = \sum_{k=0}^{H-1} \gamma^k g(\hat{o}_{t+k}, a_{t+k}) + \gamma^H V_\psi(\hat{o}_{t+H}), \tag{36}$$

where $g(\cdot)$ is the reward evaluator.

*Iterative action-sequence planning.* Following the trajectory optimization scheme used in TD-MPC2 [22], we employ an iterative sampling procedure. At each planning iteration $j$, the MPC maintains a factorized Gaussian proposal distribution over action sequences:

$$q^{(j)}(\mathbf{a}) = \prod_{k=0}^{H-1} \mathcal{N}(a_{t+k}; \mu_k^{(j)}, (\sigma_k^{(j)})^2 I). \tag{37}$$

For the initial iteration ($j = 1$), the mean sequence $\mu^{(1)}$ is initialized to the center of the valid action space. At each iteration, we sample $N$ candidate action sequences from this proposal distribution.

Each sampled sequence is rolled out with the IEWM and scored by Eq. (36). We then select the top $K$ sequences to update the distribution parameters $\mu^{(j+1)}$ and $\sigma^{(j+1)}$ for the next iteration using the Cross-Entropy Method (CEM). After $M$ planning iterations, the controller selects the highest-scoring sequence from the final elite set,

$$\mathbf{a}^\star_{t:t+H-1} = \arg\max_{\mathbf{a}^{(i)}_{t:t+H-1}} J(\mathbf{a}^{(i)}_{t:t+H-1}), \tag{38}$$

and executes only its first action $a_t^\star$. This receding-horizon procedure replans after every real environment transition.

*Hyperparameters.* We use the MPC hyperparameters detailed in Tab. 6.

**Table 6: MPC hyperparameters.**

| Hyperparameter | Value |
|---|---|
| Planning horizon $H$ | 6 |
| Number of candidate sequences $N$ | 16 |
| Number of elite samples $K$ | 4 |
| Planning iterations $M$ | 3 |
| Discount factor $\gamma$ | 0.98 |

## D.3 Online RL Policy Learning

All methods are trained for a total of $10^6$ timesteps. For the discrete action space, DQN utilizes a `Discrete` representation, while all other algorithms employ a `MultiDiscrete` action space, modeled as factorized categorical distributions over action branches. To ensure a fair comparison, model-based extensions (MBVE, MBPO) are implemented using the same PPO backbone. Detailed hyperparameters for our proposed DECARB, the model-free baselines, and the model-based baselines are summarized in Tab. 7, Tab. 8, and Tab. 9, respectively.

## D.4 Computational Cost

All experiments are conducted on an NVIDIA A100 GPU. The offline training of the indoor environmental world model requires about 12 hours. Subsequently, the downstream online RL training takes an average of 10 hours in SemiBuildingSim and 16 hours in the Sinergym environment. During inference, one time step diffusion costs about 150ms, ensuring the real-time control of HVAC systems.

**Table 7: Hyper-parameters for the proposed ADAPT ($H_{fore}, H, T$ on SemiBuildingSim).**

| Hyper-parameter | Value | Hyper-parameter | Value |
|---|---|---|---|
| Optimizer | Adam | Learning rate | $2\times10^{-3}$ |
| Discount factor $\gamma$ | 0.99 | Replay buffer size | $5\times10^{5}$ |
| Batch size | 64 | Exploration $\varepsilon$ | $1.0 \rightarrow 0.04$ |
| Target update interval | 1000 steps | TD($\lambda$) parameter | 0.8 |
| Bootstrap horizon $H$ | 3 | Forecaster horizon $H_{\text{fore}}$ | 3 |
| Network architecture | $(512, 512)$ | Forecaster History step $T$ | 6 |

**Table 8: Hyper-parameters for Model-Free Baselines.**

| Parameter | DQN | BDQ | A2C | PPO | TransformerRL |
|---|---|---|---|---|---|
| Learning rate | $2\times10^{-3}$ | $2\times10^{-3}$ | $8\times10^{-4}$ | $10^{-3}$ | $10^{-3}$ |
| Discount factor $\gamma$ | 0.99 | 0.99 | 0.98 | 0.98 | 0.98 |
| Batch size | 64 | 64 | 40 | 1200 | 1200 |
| Network hidden dim | 256 | 512 | 256 | 256 | 256 |
| GAE $\lambda$ | – | – | 0.9 | 0.8 | 0.8 |
| PPO Clip $\varepsilon$ | – | – | – | 0.2 | 0.2 |
| Entropy coef. | – | – | 0 | 0.01 | 0.01 |

**Table 9: Hyper-parameters for Model-Based Baselines (MBVE and MBPO built on the PPO backbone).**

| Parameter | MBVE | MBPO | DreamerV3 |
|---|---|---|---|
| Rollout / Imagination horizon | 8 | 8 | 8 |
| Transitions / Starts per iter. | 256 | 1024 | 256 |
| Blend / Return $\lambda$ | 1.0 | – | 0.95 |
| Warmup (iters) | 3 | 3 | 3 |